\documentclass[]{alaya}

\usepackage{makecell}
\usepackage{wrapfig}
\usepackage{tabularx}
\usepackage{textcomp}
\usepackage{stfloats}
\usepackage{url}
\usepackage{verbatim}
\usepackage{titlesec}
\usepackage{tocloft}
\usepackage{adjustbox}
\usepackage{multirow}
\usepackage{pifont}
\usepackage[sc]{mathpazo}
\usepackage{tikz}
\usetikzlibrary{positioning,fit,backgrounds,arrows.meta,calc}
\usepackage{comment}
\usepackage{amsmath,amssymb}
\usepackage{colortbl}
\usepackage[sort&compress]{natbib}
\usepackage{color}
\usepackage{booktabs}
\usepackage{hyperref}
\usepackage{graphicx}
\usepackage{subcaption}
\RequirePackage{xspace}
\makeatletter
\DeclareRobustCommand\onedot{\futurelet\@let@token\@onedot}
\def\@onedot{\ifx\@let@token.\else.\null\fi\xspace}
\usepackage[most]{tcolorbox}
\usepackage{array}
\usepackage{siunitx}
\usepackage[table]{xcolor}
\usepackage{caption}
\definecolor{headerpurple}{HTML}{d8d2fc}
\definecolor{rowgray}{gray}{0.95}
\usepackage{CJKutf8}

\makeatother

\definecolor{adptorange}{RGB}{248, 205, 172}
\definecolor{cmpblue}{RGB}{189, 215, 238}
\definecolor{our_red}{RGB}{232,157,160}
\definecolor{our_blue}{RGB}{136,206,230}
\definecolor{our_orange}{RGB}{246,200,168}
\definecolor{our_green}{RGB}{178,211,164}
\definecolor{mygray}{HTML}{f0f0f0}
\definecolor{token_blue}{RGB}{84, 120, 140}

\usepackage{bbding}
\usepackage{fontawesome}
\usepackage{float}

\newlength\savewidth
\newcommand{\tablestyle}[2]{\setlength{\tabcolsep}{#1}\renewcommand{\arraystretch}{#2}\centering\footnotesize}
\newcolumntype{x}[1]{>{\centering\arraybackslash}p{#1pt}}
\newcolumntype{y}[1]{>{\raggedright\arraybackslash}p{#1pt}}
\newcolumntype{z}[1]{>{\raggedleft\arraybackslash}p{#1pt}}

\renewcommand{\paragraph}[1]{\vspace{1.25mm}\noindent\textbf{#1}}

\usepackage{algorithm}
\usepackage{listings}
\definecolor{codeblue}{rgb}{0.25, 0.5, 0.5}
\definecolor{codekw}{rgb}{0.35, 0.35, 0.75}
\lstdefinestyle{json}{
    basicstyle = \fontsize{7.4pt}{8.4pt}\selectfont\ttfamily,
    columns = fullflexible,
    aboveskip=2pt, belowskip=2pt,
    breaklines = true,
    showstringspaces=false,
    commentstyle = \color{codeblue},
    keywordstyle = \color{codekw},
}

\newcommand{\method}{\texttt{WorldRover}\xspace}
\newcommand{\engine}{\texttt{WorldRover-Engine}\xspace}
\newcommand{\data}{\texttt{WorldRover-10M}\xspace}

\definecolor{boxgrey}{HTML}{808080}
\definecolor{boxred}{HTML}{CD3B3A}
\definecolor{boxgreen}{HTML}{6F9C5E}
\definecolor{boxblue}{HTML}{4D8CC2}
\definecolor{boxyellow}{HTML}{FFFD78}
\definecolor{boxorange}{HTML}{FFC000}
\definecolor{trajOrange}{HTML}{F5871F}
\definecolor{trajGreen}{HTML}{4A7E49}
\definecolor{trajBlue}{HTML}{2B8FE6}
\definecolor{trajRed}{HTML}{FF2020}
\definecolor{trajPurple}{HTML}{9A5CFF}

\title{
WorldRover: A Scalable Synthetic Video Data Engine for World Exploration with Rich Annotations
}

\author[1,2*]{Xiaojie Xu}
\author[1,2*]{Zhengyuan Lin}
\author[1,2]{Runyi Li}
\author[1]{Yihao Liu}
\author[1\dagger]{Kaipeng Zhang}
\author[1\dagger]{Yongtao Ge}

\affiliation[1]{Alaya Lab}
\affiliation[2]{The University of Tokyo}

\abstract{
Learning to generate or reconstruct explorable worlds requires video paired with more than RGB:
camera motion, scene geometry, temporal correspondence and, for interactive models, control
signals. Real capture can provide some of these signals, but dense geometry and long-range
correspondence usually rely on estimation or specialised instrumentation. Rendering provides these
quantities directly, yet existing synthetic resources rarely combine them on the same frames while
also supporting controlled changes of viewpoint and appearance. We introduce \method, a data
engine for generating richly annotated, long-range explorations of artist-built environments. At its
core, \engine is an Unreal Engine pipeline that executes and offline-renders minute-
scale routes while preserving their full trajectories and scene geometry. The same exploration
can be replayed from first-person, third-person, and $360^\circ$ panoramic cameras under different
environmental states. Using \engine, we construct \data, whose sequences
pair RGB with metric depth, camera trajectories, and trajectory-derived action signals throughout
each exploration. Third-person subsets additionally provide dense optical flow, long-range 2D/3D
point tracks with visibility, and a character trajectory distinct from the camera trajectory. The
engine can render a traversal from first-person, third-person and $360^\circ$ panoramic viewpoints,
under different environmental states or with a neutral white material, while preserving the route
and scene geometry. \method therefore turns long-horizon world exploration into a scalable
data-generation problem, providing supervision for models that must build, maintain, and revisit
coherent representations of an explorable world.
}

\github{\url{https://alayalab.github.io/WorldRover}}
\Code{\url{https://github.com/AlayaLab/WorldRover}}
\metadata[Data]{\url{https://huggingface.co/datasets/xjxu21/WorldRover}}
  
\date{August 15, 2026}

\begin{document}
\maketitle
\footnotetext[1]{$^{*}$Equal contribution}
\footnotetext[2]{$^{\dagger}$Corresponding authors: \texttt{kaipeng.zhang@shanda.com}, \texttt{yongtao.ge@shanda.com}}

\begin{figure}[p]
    \centering
    \includegraphics[width=\linewidth]{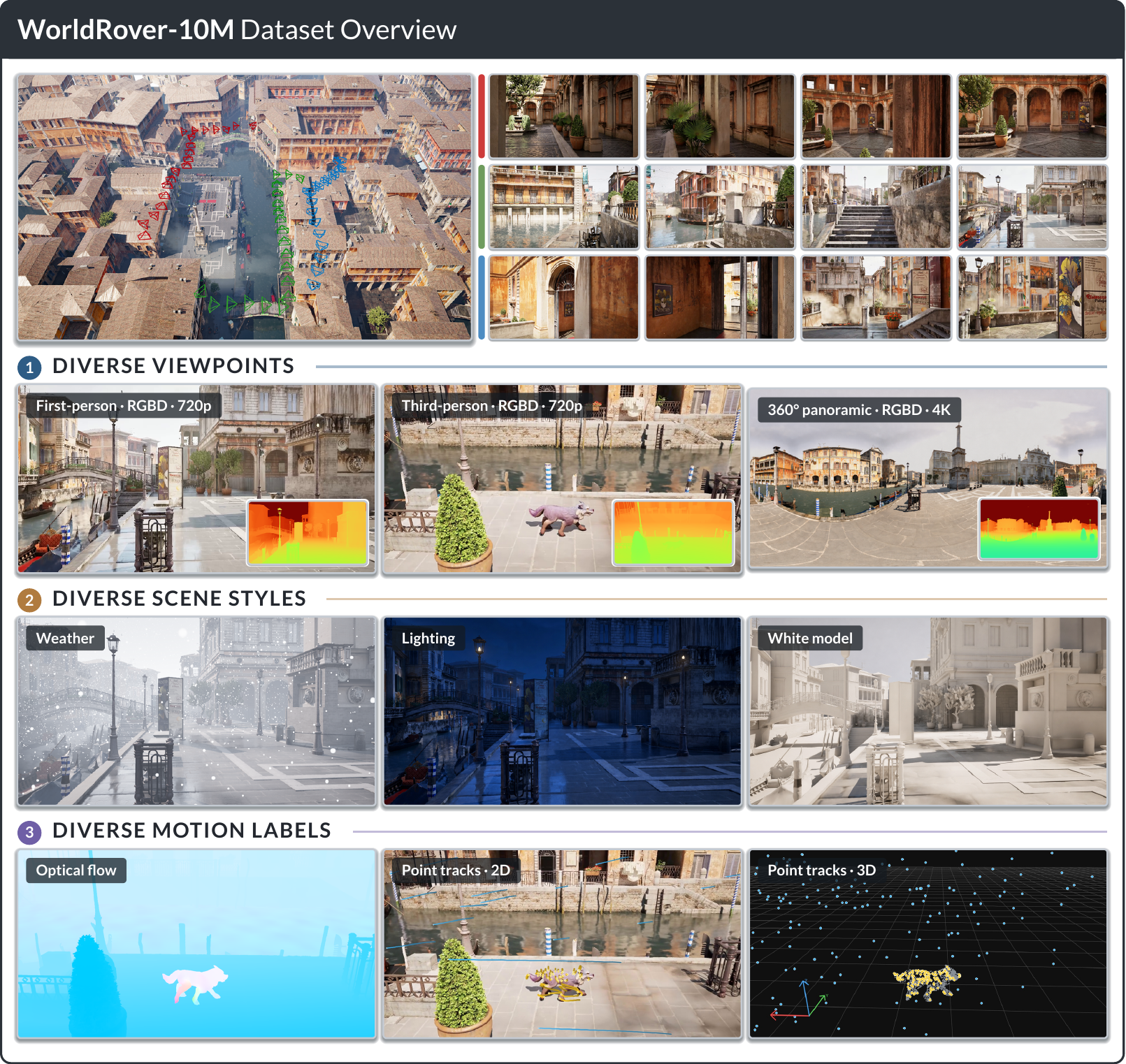}
    \vspace{2pt}
    \caption{\textbf{\data} at a glance. Four rows, top to bottom.
    \emph{Scene and routes}: one UE environment with three routes drawn as camera frustums, and four
    frames along each, colour-keyed to the overview.
    \emph{(1) Diverse viewpoints}: the same instants observed as first-person, third-person and
    $360^\circ$ panoramic video; every panel is RGB with metric depth inset, released with the
    camera that produced it. Third-person adds the character's own trajectory and the
    trajectory-derived action stream.
    \emph{(2) Diverse scene styles}: the first-person view re-rendered under a different weather
    state, a different lighting state, and as an untextured white model, with scene, motion and
    camera held fixed.
    \emph{(3) Diverse motion labels}: dense optical flow and long-range point tracks, shown in 2D
    and in world space; a track stores a 3D position, its 2D projection and visibility per frame.
    Third-person only, with independent body and camera motion.}
    \label{fig:teaser}
\end{figure}

\section{Introduction}
\label{sec:intro}

An explorable visual world must remain coherent as the observer moves through it. As the viewpoint
changes or an action moves an agent, a model should preserve scene layout, maintain object identity
through occlusion and distinguish camera motion from changes in the world. These requirements
appear in interactive world generation~\citep{agarwal2026cosmos,ball2025genie3,alayaworld2026,xu2026worldmark,mao2026yume1, gao2026infinite} and in dynamic 4D reconstruction~\citep{omnix2026}. Although the two settings
have different outputs, they share a data problem: RGB alone leaves the causes of visual change
entangled. Camera motion and object motion both displace pixels, illumination can change without
geometry changing, and an occluded surface has no visible correspondence target. Camera
trajectories, metric geometry, temporal correspondence and, when available, action signals make
these factors explicit.

Data with this combination is difficult to collect. Large real-video corpora provide broad visual
coverage~\citep{sekai2025}, but camera pose, depth and correspondence are generally estimated after
capture; pose can drift, correspondence can fail through occlusion, and instrumented capture
restricts the environments and conditions that can be recorded. Rendering exposes these quantities
as part of scene state and image formation, but existing synthetic resources are optimized for
different goals. Task-specific datasets provide dense labels for a fixed
benchmark~\citep{butler2012naturalistic,roberts2021hypersim,wang2020tartanair,zhou2025omniworld};
interactive simulators prioritize low-latency observations for a
policy~\citep{shah2018airsim,dosovitskiy2017carla}; and procedural generators
increase the number of scenes~\citep{greff2022kubric,raistrick2023infinite}. Across these forms,
the same world traversal is seldom re-observed under controlled changes of viewpoint and
appearance. The missing capability is therefore not merely more rendered frames, but a production
system that scales variation while preserving the factors shared across related observations.

We introduce \method, a synthetic video data engine built around one abstraction: an exploration
is a reusable world-space trajectory, not a single finalized video. \engine prepares an
artist-built UE environment once, generates minute-scale routes through it and renders those routes
offline. The same exploration can then be observed from first-person, third-person and
$360^\circ$ panoramic cameras, or re-rendered under another environmental state or as a neutral
white model. Because route, timing and scene structure are retained across these variants, the
resulting pairs isolate viewpoint or appearance changes instead of confounding them with a
different motion. Camera records and renderer-derived annotations remain tied to the frames they
describe.

Using \engine, we construct \data: 6,003 sequences from 32 environments, comprising 21.9M rendered
frames, 202.7 hours of video and 18.7\,TB of released data. The name refers to its 10.8M
first-person frames; third-person and panoramic subsets provide additional observations of the
exploration corpus. Annotation bundles are task-specific rather than universal. Depending on the
subset, a sequence pairs RGB with metric depth, dense optical flow, long-range 2D/3D point tracks,
rendered camera and character trajectories, and trajectory-derived actions. This organization
supports both broad training data and controlled comparisons in which geometry, motion, viewpoint
or appearance is held fixed.

Our contributions are:
\begin{itemize}
\itemsep1pt
\item \textbf{A scalable data engine for world exploration.} \engine combines one-time scene
preparation, automated and recorded route generation, offline rendering, streaming annotation
extraction and parallel campaign execution over artist-built UE environments.
\item \textbf{Controlled re-observation of a shared trajectory.} \method renders matched
first-person, third-person and panoramic observations, as well as environmental and white-model
variants, while retaining the underlying route, timing and scene structure.
\item \textbf{A large, richly annotated release.} \data provides 21.9M frames across long
explorations, with subset-specific geometry, motion, correspondence, camera, character and action
annotations under a common release schema.
\end{itemize}

\begin{figure}[htbp]
    \centering
    \includegraphics[width=0.8\linewidth]{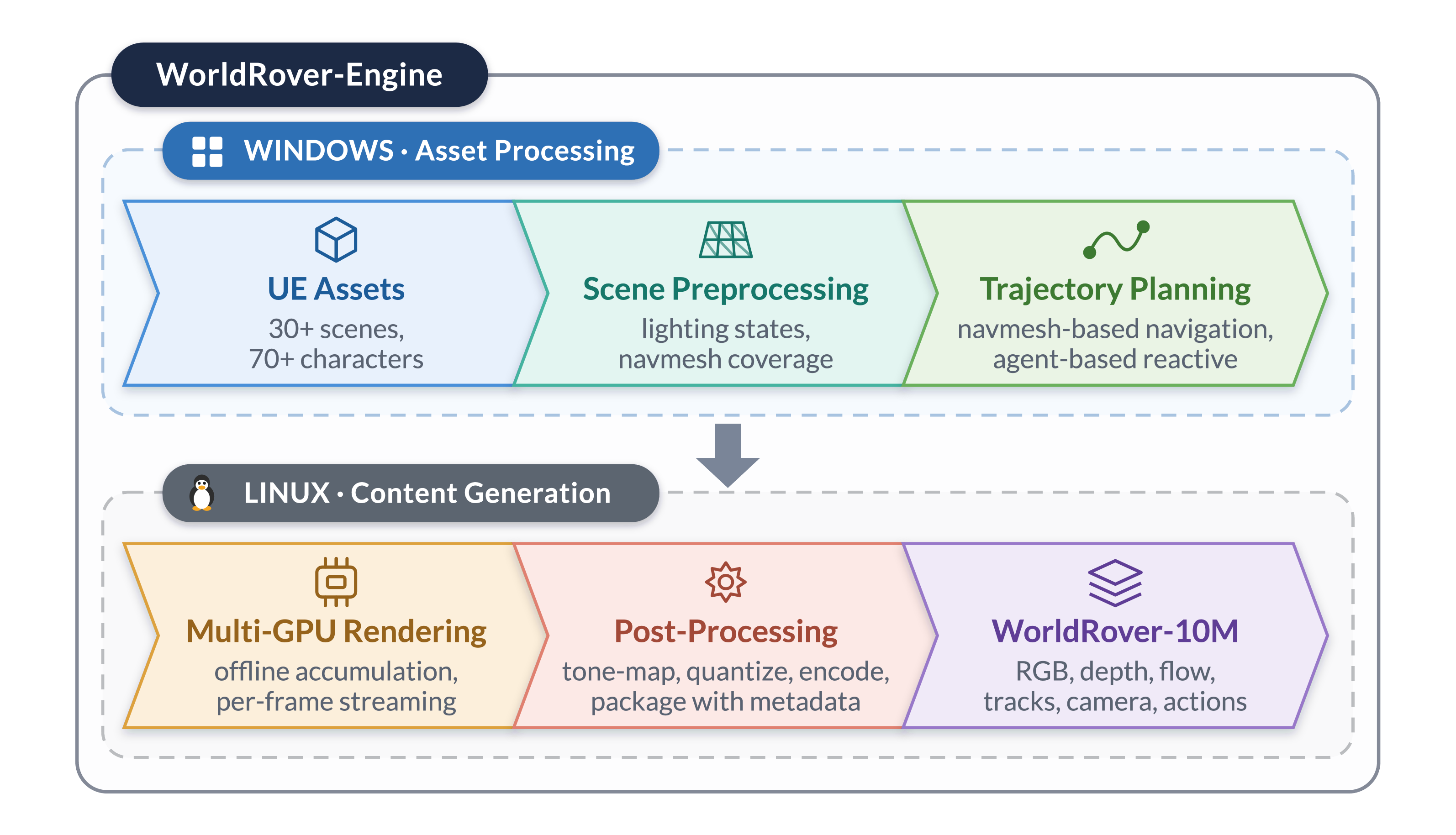}
    \caption{\textbf{\engine overview.} Asset processing runs once per scene on a Windows
    editor host; content generation runs once per sequence, headless, on a multi-GPU Linux host.
    The two halves meet at a versioned on-disk hand-off, so a campaign is replayable without an
    editor session.}
    \label{fig:engine}
\end{figure}

\section{WorldRover-Engine}
\label{sec:engine}\label{sec:overview}

\subsection{Overview}

A \method sequence follows a route through a prepared environment. Sparse route samples are
converted to human-scale camera motion with approximately constant travel speed, smoothed and
rate-limited heading, and occasional \emph{dwells} in which translation stops while the view
rotates. The resulting trajectory is then rendered from the viewpoint specified by the target
subset.

\subsection{Editor side}

\paragraph{Asset ingest.} Each scene enters as an Unreal asset pack and is prepared once on the
editor host into a uniform, headless-loadable form, allowing render workers to open it without an
editor session. Packs target a range of engine versions, so ingest first resaves each scene for
UE~5.5. Packs may also split content across streaming sub-levels or World Partition cells.
Ingest loads these components before navigation baking and stylization, and the render configuration
keeps the required cells resident during headless generation. A navigation mesh is then baked over
the walkable surface and stored with the scene for the planner.

\paragraph{Scene stylization.} Each appearance state of \Cref{sec:lightingstates} is authored in the
editor so that a headless render reproduces it with no runtime setup. Its visual configuration is
prepared there, while exposure is calibrated through the offline rendering path.

\begin{itemize}
\itemsep2pt
\item \textbf{Environmental states.} A state is authored entirely on the scene side and saved as
its own map. It replaces the scene's shipped lighting with a fixed rig in the persistent level ---
a directional sun, a sky light for ambient fill, a sky-atmosphere and exponential height fog whose
density and in-scattering colour define the atmospheric state --- and may add rain, snow or drifting
mist through persistent Niagara systems that cover the walkable area. The state does not depend on
trigger volumes or level-blueprint events, so it is active when the map loads in a headless session.
A campaign selects the corresponding map variant rather than configuring these effects per shot.

\item \textbf{Fixed exposure.} Each environmental state uses its own calibrated fixed exposure, and
auto-exposure remains disabled during rendering. A night state can therefore remain dark rather
than adapting toward mid-grey. The value for a state is calibrated once: a probe frame is rendered
through the same offline path used by a campaign, its mean luminance is measured, and the exposure
in the map's post-process volume is adjusted until the frame reaches the intended target. All
sequences rendered from that state then use the calibrated value without per-shot adjustment.
\end{itemize}

\begin{figure}[htbp]
    \centering
    \includegraphics[width=\linewidth]{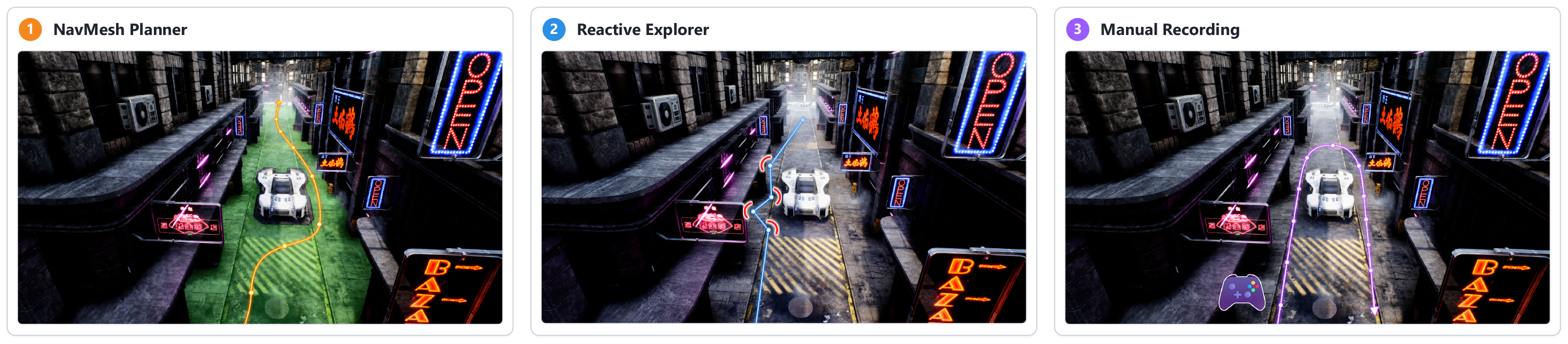}
    \caption{\textbf{Trajectory generation.} The three strategies for producing walking
    trajectories, on one scene. \textbf{(1)~\textcolor{trajOrange}{NavMesh planner}:} a route
    (\textcolor{trajOrange}{orange}) planned over a baked navigation mesh
    (\textcolor{trajGreen}{green}); the mesh is built once and reused, so planning is fast and
    coverage broad, at the cost of per-scene tuning. \textbf{(2)~\textcolor{trajBlue}{Reactive
    explorer}:} a path (\textcolor{trajBlue}{blue}) driven by local obstacle sensing
    (\textcolor{trajRed}{red}); it needs no precomputed map, but senses step by step and can walk
    itself into a corner. \textbf{(3)~\textcolor{trajPurple}{Manual recording}:} a human-driven
    path (\textcolor{trajPurple}{purple}), which yields a specific trajectory on demand at the cost
    of manual effort. Together they trade off automation, coverage and control.}
    \label{fig:traj}
\end{figure}

\paragraph{Trajectory generation.}
\label{sec:traj}
We produce walking trajectories with a navigation-mesh (NavMesh) planner, a reactive explorer and
manual recording (\Cref{fig:traj}). The three methods trade off automated coverage, dependence on
scene preparation and direct control over the route.

\begin{itemize}
\itemsep2pt
\item \textbf{NavMesh planner.} An agent starts at a random reachable point and walks toward sampled
goals on the baked navigation mesh, with occasional changes of heading. The resulting routes remain
on the walkable surface and tend to follow efficient connections through the scene.

\item \textbf{Reactive explorer.} A physics-driven agent uses a fan of short ray casts to avoid
nearby obstacles and favours directions it has not recently visited. It requires no precomputed
navigation mesh, but its local policy may wander, backtrack or enter a dead end.

\item \textbf{Manual recording.} An operator walks through the scene while the
camera path is recorded. This provides routes and viewpoints selected deliberately rather than by
an automatic exploration policy.
\end{itemize}

\subsection{Server side}

\paragraph{Offline rendering and dynamic illumination.} Frames are produced through UE's Movie
Render Queue, which accumulates spatial sub-samples for each output frame. The pipeline targets
batch production rather than interactive latency, allowing more rendering work per frame; in our
configuration, accumulation reduces aliasing and temporal variation in Lumen illumination relative
to the single-sample interactive path. Lighting is evaluated at render time with Lumen rather than
read from a baked lightmap. Re-rendering an authored environmental state therefore updates direct
and indirect illumination, including colour bleeding and shadow softness, rather than applying an
image-space recolouring.

\paragraph{Sequence construction and streaming post-processing.} SPEAR\cite{spear2026} provides the control plane
used to load and configure the UE project, create the required actors and submit render jobs. Sparse
waypoints are first converted to dense per-frame keyframes by constant-speed reparameterization,
position and heading smoothing, and yaw-rate limiting. The keyframes are written to a Level
Sequence consumed by Movie Render Queue; for panoramic runs, the six cube-face orientations are
assigned at this stage. Movie Render Queue writes multi-layer EXR frames, while host-side workers
perform tone mapping, depth extraction and, for panoramic sequences, cube-face stitching. Because
these operations depend on one rendered instant at a time, each EXR is processed and deleted after
it appears rather than retained until the sequence finishes, keeping intermediate storage bounded.
The worker pool is sized so that its sustained consumption rate remains above the renderers'
production rate.

\paragraph{Parallel scheduling and recovery.} Throughput was limited mainly by UE game-thread work
and host-side post-processing; one render instance used only
a fraction of a GPU. Multiple instances therefore share each GPU subject to video-memory limits,
and RGB is encoded on the dedicated hardware encoder so CPU cores remain available for EXR
processing. Render instances are reused across several sequences to amortize startup, shader
compilation and lighting warm-up. A watchdog terminates instances that do not exit cleanly after a
batch.

\section{WorldRover-10M: Three Camera Viewpoints}
\label{sec:views}\label{sec:cameras}

The pipeline defines one first-person camera trajectory per route (\Cref{sec:engine}), and the
other viewpoints are derived from that trajectory. In third-person, an animated character
walks the recorded path and a follow camera holds a fixed offset behind and above it, so the
character trajectory of a third-person sequence is the camera trajectory of the corresponding
first-person sequence. In panoramic mode the camera follows the recorded path and only
frame assembly differs: six differently oriented cameras are rendered at each instant and stitched.
This construction provides three observations of a route with matched timing, scene geometry and
environmental state.

\subsection{First-person}
An eye-height camera rides the trajectory itself, with pitch and roll locked and heading following
the direction of travel, smoothed and rate-limited so the motion reads as human locomotion rather
than as a planner's output. Here the camera \emph{is} the agent, so the camera trajectory is also
the agent trajectory, and the trajectory-derived action stream, pose and imagery share the same
motion. This is the mode that matches egocentric video and embodied navigation.

\subsection{Third-person}
\begin{figure}[htbp]
    \centering
    \includegraphics[width=0.5\linewidth]{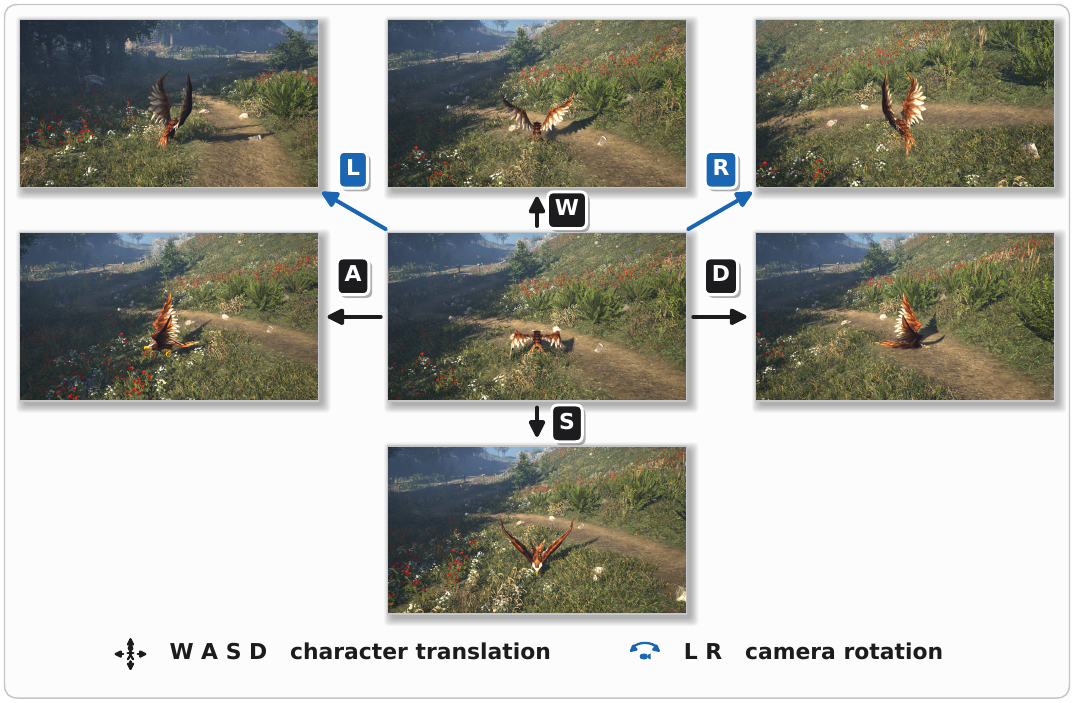}
    \caption{\textbf{Third-person control semantics.} Centre: the start state. \textbf{W/A/S/D}
    (black) translate the character along camera-relative directions --- W away from the camera, S
    toward it, A and D to its left and right --- and the character turns to face its heading while
    the camera follows with a slowly varying orbit azimuth. \textbf{L/R} (blue) yaw the camera
    about a stationary character, so the character holds its image position while the
    scene rotates around it.}
    \label{fig:thirdperson}
\end{figure}

An animated character replays the recorded trajectory in world coordinates, and a follow camera
maintains a nominal offset behind and above it. Camera azimuth is driven by a slowly varying orbit
target rather than copied from the character heading, so the view remains approximately stable
through ordinary turns while still following long changes in travel direction. Character motion can therefore appear toward,
away from or across the camera, and the trajectory-derived action stream is expressed in this
camera-relative frame
(\Cref{fig:thirdperson}). The character turns toward its direction of travel, with a rate limit,
while explicit camera-yaw controls rotate the view around a stationary character. The camera and
character trajectories are distinct, although the camera trajectory is derived from the character
path. The same nominal camera rig is used across characters.

These sequences contain an articulated subject observed from a moving camera, with separate
recorded trajectories for the subject and camera. This separation supports evaluation of methods
that must distinguish camera motion from non-rigid subject motion.

\subsection{\texorpdfstring{$360^\circ$}{360-degree} panoramic}
\begin{figure}[htbp]
    \centering
    \includegraphics[width=\linewidth]{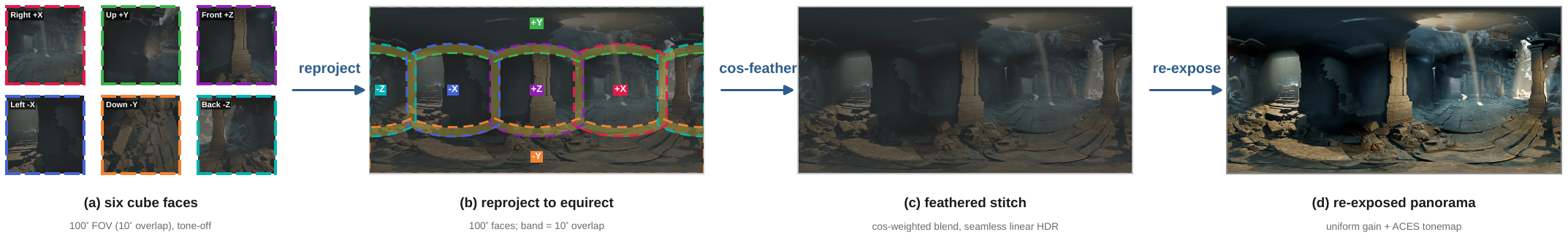}
    \caption{\textbf{Panoramic assembly.} (a) Six cube faces at $100^\circ$ FOV, giving a
    $10^\circ$ overlap between neighbours, rendered with tone mapping disabled so all faces share
    one radiometric scale. (b) Reprojection onto an equirectangular image; dashed outlines,
    colour-matched to (a), mark each face's region and the yellow bands its overlap, which exposes
    the pincushion distortion of the side faces and the polar caps. (c) A cos-weighted feather blend
    merges the overlaps into a linear-HDR panorama. (d) Exposure is estimated with
    latitude-aware luminance weighting and smoothed across frames; local base/detail tone mapping
    compresses the result before sRGB conversion.}
    \label{fig:panoassembly}
\end{figure}

Equirectangular video at $4096\times2048$ is assembled from six cube faces rendered at the same
instant (\Cref{fig:panoassembly}), providing an omnidirectional observation around the camera.
We do not use UE's own panoramic pass, which hard-clamps scene colour during accumulation: in a
tone-mapping-off render the brightest tenth of the image collapses onto a single value, and those
highlights cannot be recovered afterwards. We instead render six ordinary perspective faces, which
retain the measured HDR range, and assemble the panorama on the host. The faces are reprojected and
cosine-feathered in linear light. Exposure is estimated from a cosine-latitude-weighted
log-luminance statistic and smoothed over time for video; a large-support bilateral base/detail
operator then compresses the dynamic range before sRGB conversion. Panoramic and first-person
modes share camera centres and orientations, allowing a perspective view to be compared with a
perspective reprojection of the panorama.

\section{WorldRover-10M: Multimodal Data and Annotations}
\label{sec:modalities}\label{sec:gt}

A sequence is delivered as a directory with a manifest that lists the available outputs. Annotation
bundles differ by subset, but use common coordinate, timing and metadata conventions.

\paragraph{One EXR frame, three registered rasters.} Movie Render Queue renders each output frame
through a single configuration that writes colour, a depth layer and a velocity layer into one
multi-layer EXR, accumulating two spatial samples per frame for anti-aliasing. Colour, depth and
screen-space motion are therefore the same rasterization of the same instant, at the same
resolution and projection; the camera annotation is the camera that produced that frame. Reading
those layers is a purely intra-frame operation --- each is extracted and quantized without
reference to neighbouring frames. This permits the streaming process described in
\Cref{sec:engine}.

\subsection{RGB video}
Photorealistic colour video under dynamic global illumination, rendered offline with sub-sample
accumulation to reduce aliasing and temporal variation in indirect light, distant lights and specular
highlights. Frames are tone-mapped from the linear EXR and encoded to H.264 on the GPU's dedicated
encoder block, keeping CPU cores available for per-frame extraction and panorama stitching. First-
and third-person videos are $1280\times720$;
panoramic video is $4096\times2048$ equirectangular, assembled from six $1536^2$ cube faces.

\subsection{Metric depth}
\label{sec:depth}
Depth is read from the renderer's depth layer, which stores radial distance from the camera centre
in metres. Unlike an estimated or sensed depth map, it does not depend on image texture or sensor
noise; the released representation nevertheless has a fixed valid range. Values are clipped to
$0.1$--$200$\,m and quantized logarithmically into 16 bits,
\[
t \;=\; \frac{\log d - \log d_{\mathrm{near}}}{\log d_{\mathrm{far}} - \log d_{\mathrm{near}}},
\qquad d_{\mathrm{near}} = 0.1\,\mathrm{m},\quad d_{\mathrm{far}} = 200\,\mathrm{m},
\]
and the frames are encoded losslessly with FFV1. A sidecar carries the inverse formula and encoding
parameters. The logarithmic parameterization gives finer absolute precision at short range and
coarser precision at longer distances.

\subsection{Optical flow}
Third-person subsets provide dense optical flow derived from the renderer rather than estimated
from the images. A post-process material bound to an extra
render pass writes the engine's own velocity buffer into a named sublayer of the same multi-layer
EXR that carries colour and depth, as a float16 buffer whose first two channels hold the
screen-space displacement $(\Delta u, \Delta v)$ of each pixel between consecutive frames,
normalized so that the full image width spans $2$. Pixel displacements follow from the image size,
\[
  \Delta u_{\mathrm{px}} = \Delta u \cdot \tfrac{W}{2}, \qquad
  \Delta v_{\mathrm{px}} = \Delta v \cdot \tfrac{H}{2},
\]
with $+v$ pointing down the image. Because temporal anti-aliasing contaminates the velocity buffer,
it is disabled in this render configuration and anti-aliasing is spatial instead --- the accumulated
sub-samples of \Cref{sec:engine} --- so the released flow is the raw per-pixel motion of the frame
it accompanies. Its value does not depend on texture or image matching, but it follows the engine's
velocity-buffer semantics and the storage range described below.

The field is released as one lossless FFV1 stream per sequence, rather than one file per frame.
Each frame is mapped affinely onto two 16-bit channels with a single scale $g$, in pixels per
frame, that is fixed for the whole release ($64$ by default and configurable per campaign),
\begin{align*}
  u_{16} &= \mathrm{clip}\!\left[\tfrac{65535}{2}\left(\Delta u_{\mathrm{px}}/g + 1\right)\right],\\
  \Delta u_{\mathrm{px}} &= \left(2\,u_{16}/65535 - 1\right)g,
\end{align*}
and likewise for $v$; displacements beyond $\pm g$ are clipped. Thus the released stream preserves
the rendered motion field within $\pm64$ pixels per frame under the default configuration, while
larger displacements saturate unless a campaign uses a larger scale. A sidecar records $g$, the frame
count, resolution, frame rate and the decode formula, so the stream decodes from the released files
alone. Lossless video encoding exploits both the spatial smoothness and temporal redundancy of the
field, avoiding the overhead of one file per frame. The payload is 16-bit and must be read with a
decoder that preserves that depth.

\subsection{Long-range point tracks}
\label{sec:tracks}

\begin{figure}[H]
    \centering
    \includegraphics[width=\linewidth]{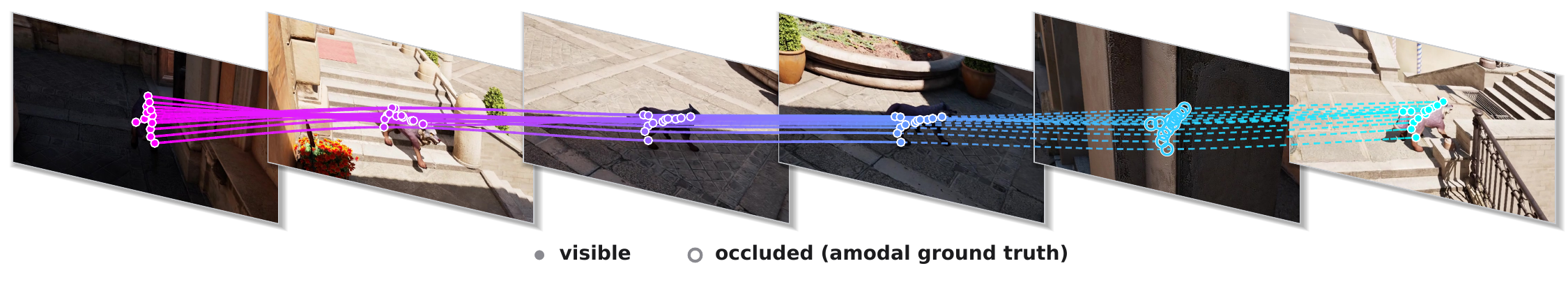}
    \caption{\textbf{Point-track ground truth.} Six frames $4$\,s apart, time encoded magenta
    $\rightarrow$ cyan; each curve follows one 3D point on the character. Filled markers and solid
    curves are visible, hollow rings and dashed curves occluded and drawn at their amodal
    ground-truth position. In the fifth panel a wall encloses the camera and hides the character
    completely, yet the tracks retain their amodal positions with zero visibility flags.}
    \label{fig:tracks}
\end{figure}

Third-person sequences carry long-range point tracks that are geometric rather than matched. On a
seed frame a batch of query pixels is unprojected into world coordinates using the rendered depth
and the camera of that frame; on later frames those world points are projected back into the
image, and the resulting pixel sequence is the track. New cohorts are seeded periodically, so
surfaces revealed later still receive query points, and character and background pixels are drawn
separately. Because propagation uses projection rather than image matching, it does not accumulate
matching drift; accuracy instead depends on the rendered depth, camera pose and the binding of a
point to the character.

Background points keep the world coordinate obtained at their seed frame. The default character
tracker stores the seed point in the character's local frame and transforms it by the per-frame root
position and yaw. This rigid binding follows the character trajectory but not articulated
deformation, so points on a foot, tail or head can move away from the rendered surface during an
animation cycle. We also implement an LBS variant for sequences whose skinned mesh can be
reconstructed from additional glTF skinning weights and per-frame bones. It binds each query to the
reconstructed skinned surface and therefore follows its deformation; those mesh and skeleton inputs
are not part of the sequence archive.

Occlusion does not interrupt a track. Position is written unconditionally, and visibility is a
separate per-frame flag, set by comparing a track's projected depth against the rendered depth at
that pixel. The 2D trajectory is therefore \emph{amodal}: while a point is hidden, the release still
states where in the image it would be (\Cref{fig:tracks}). Visibility and having a defined position
are separate, as in PointOdyssey's occlusion labels~\citep{zheng2023pointodyssey}; conflating them
removes supervision during occlusion.

The rigid archive stores the seed, per-frame world and image positions, visibility, object ID and
initial binding residual. The LBS archive omits world positions and binding residuals because its
character positions come from the skinned mesh. Queries can be seeded only on visible surfaces,
biasing follow-camera samples toward the body's back; background tracks also leave the frame as the
camera moves, so the longest tracks are mainly on characters.

\subsection{Camera, character and action streams}
\label{sec:trajfiles}\label{sec:actions}
The remaining streams are not rasters but per-frame records derived from the trajectories the
sequence was rendered along.

The \textbf{camera trajectory} gives the position, orientation and intrinsics of the camera at
every frame. It is written from a sidecar emitted at render time, so it describes the camera that
was actually rendered rather than the keyframes sent in. Panoramic sequences declare an
equirectangular projection and keep the pose of the shared camera centre, instead of reporting a
perspective intrinsic matrix that does not describe the image.

The \textbf{character trajectory} gives the position, orientation and speed of the character at
every frame, and third-person sequences release it alongside the camera because the two are
genuinely different motions (\Cref{sec:views}): they differ in path, and far more in heading, since
the character turns while the follow camera holds its azimuth.

The \textbf{trajectory-derived action stream} is a control representation reconstructed from the
motion rather than a log of original input events. Velocities are differenced between adjacent
trajectory frames, resolved into forward, lateral, turning and vertical axes in the corresponding
first-person camera frame, quantized, and written when an axis value changes. Third-person sequences
retain this camera-relative stream rather than recomputing actions from the follow camera.

\subsection{Release details}
\label{sec:extensibility}\label{sec:format}\label{sec:validation}

\begin{table}[htbp]
\centering
\tablestyle{5pt}{1.2}
\caption{\textbf{Released data format.} How each item is stored when a subset includes it.}
\label{tab:dataformat}
\begin{tabular}{ll}
\toprule
\textbf{Data, when present} & \textbf{Contents} \\
\midrule
RGB video                & H.264, hardware-encoded \\
Depth                    & 16-bit log-quantized, FFV1 lossless, $+$ decode sidecar \\
Optical flow             & 16-bit affine-quantized, FFV1 lossless, $+$ decode sidecar (third-person only) \\
Point tracks             & per-track 3D position, 2D projection and visibility, per frame (third-person only) \\
Camera trajectory        & per-frame pose and intrinsics of the rendered camera \\
Character trajectory     & per-frame pose and speed (third-person only) \\
Trajectory-derived actions & quantized axis events $+$ binding snapshot \\
Description              & environment, camera, map variant and render metadata \\
Route preview            & plan view of the trajectory \\
\bottomrule
\end{tabular}
\end{table}

Each sequence directory holds the encoded video, the depth stream with
its decode sidecar, the trajectory tables, the trajectory-derived action stream and a description
manifest, plus a
rendered plan view of the route for inspection. \Cref{tab:dataformat} lists how each item is stored
when a subset includes it. The description manifest records the environment, the camera and sensor
geometry, the projection model, the selected map variant, the trajectory source, sequence timing and
the render configuration. The current schema does not expose normalized
lighting and weather fields; appearance variants are identified through their map configuration.
Two screens run after rendering. The first is
format: the encoded video and each annotation stream must be present, decodable and of agreeing
length, and the depth sidecar must carry the parameters its decoding needs. The second is quality,
where files can be structurally valid while their content is unsuitable --- miscalibrated exposure
or too dark, a camera that has passed through geometry, a route that keeps revisiting the same part
of a scene, a render that has come out degraded. Sequences that fail either screen are dropped
rather than released.

\clearpage
\section{WorldRover-10M: Diversity and Scale}
\label{sec:scenes}

Artist-built environments provide authored geometry, materials and scene layout, but are less
numerous than procedurally generated scenes. \engine expands a fixed asset library by composing
scenes, environmental states, viewpoints, characters and routes. Preparing an environment is a
one-time authoring step; subsequent sequences are generated from campaign configurations and
rendering compute rather than per-frame manual annotation.

\subsection{Scenes}

\begin{figure}[H]
    \centering
    \includegraphics[width=\linewidth]{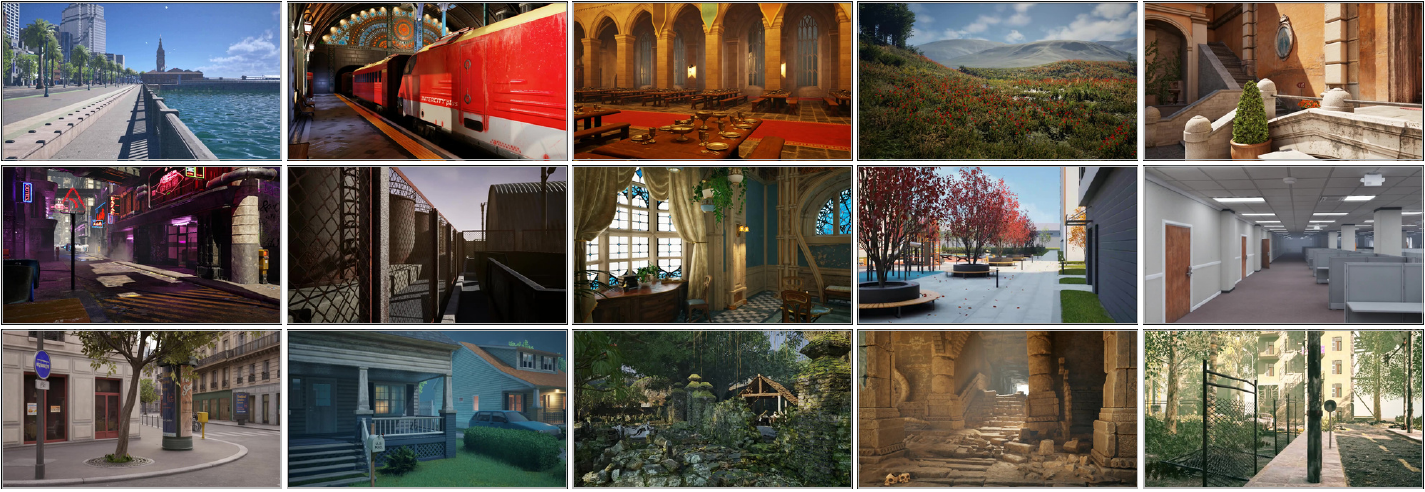}
    \caption{\textbf{A sample of the \method scene library.} Fifteen of 30+ UE
    environments --- waterfront city, transit interior, cathedral, alpine meadow, cyberpunk
    street, office interior, Paris street, suburban home, jungle ruin, and more --- all
    rendered under dynamic Lumen global illumination.}
    \label{fig:scenes}
\end{figure}

The \engine scene library comprises 30+ UE environments spanning city, historical, horror,
industrial, interior, fantasy, cyberpunk, mountain, street and town/village categories, covering
everything from a single apartment to a full urban district (\Cref{fig:scenes}). They are commercially licensed,
artist-built environments intended for games and previsualization, with authored materials, set
dressing and scene clutter.

Interior and city-scale scenes differ substantially in depth range, the density of independently
moving content and the amount of occlusion. These factors affect the difficulty of depth estimation,
optical flow and point tracking.

\subsection{Appearances}

\begin{figure}[H]
    \centering
    \includegraphics[width=\linewidth]{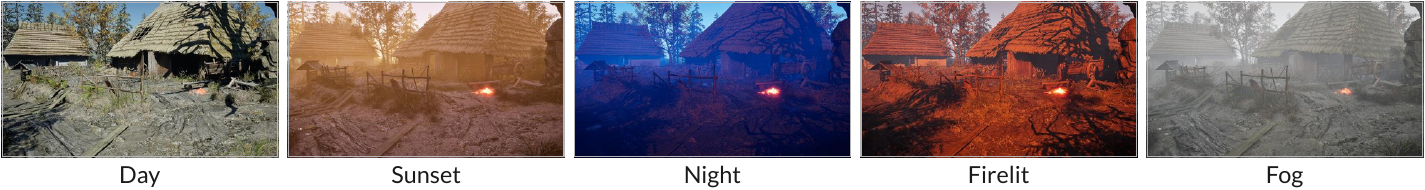}
    \caption{\textbf{Diverse appearances.} The same first-person frame of \emph{med\_village},
    rendered from an identical camera at the same instant under five environmental states. Because
    lighting is \emph{simulated at render time} rather than baked, indirect illumination, colour
    bleeding and shadow softness respond correctly to each state; each state uses its own calibrated
    fixed exposure, with no automatic adjustment during rendering, so \textbf{Night} stays dark rather
    than normalised toward mid-grey. Re-rendering
    an instant under another state therefore changes the image physically, not by recolouring a
    fixed frame.}
    \label{fig:styles}
\end{figure}
\label{sec:lightingstates}
\engine supports two appearance controls. \textbf{Lighting} sets the time of day ---
day, dawn, evening, sunset and night --- and \textbf{weather} sets the atmosphere --- clear,
overcast, fog and snow. Because lighting is simulated rather than baked, indirect illumination,
colour bleeding and shadow softness change with the authored state instead of remaining fixed in a
lightmap. Each state uses its own calibrated fixed exposure, with auto-exposure disabled during
rendering.

When a route is rendered under more than one state, geometry and motion remain fixed while
illumination changes. Such pairs can be used to separate appearance sensitivity from changes in
scene structure or trajectory.

Each state is authored as a self-contained level variant whose lighting rig is baked into the
persistent level with no gameplay-gated triggers. Its fixed exposure is calibrated through the
offline rendering path and is not adjusted automatically during rendering (\Cref{sec:engine}).

\begin{figure}[htbp]
    \centering
    \includegraphics[width=\linewidth]{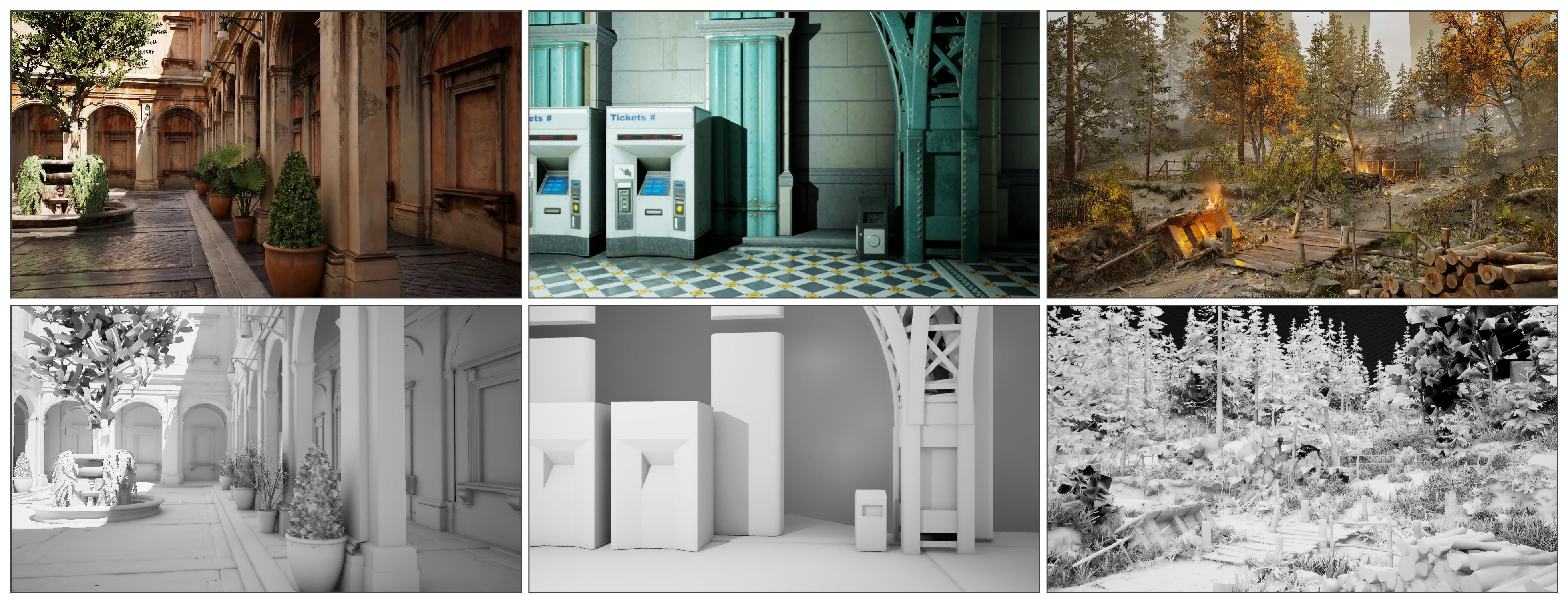}
    \caption{\textbf{White model.} Three environments, the ordinary render above and the white-model
    re-render of the same frame below. With scene surfaces drawn in the engine's default
    material, geometry and occlusion remain visible from large structures to fine detail --- arcade columns and
    planters, station ironwork, individual leaves and branches --- while texture, colour, decals and
    emissive light do not: the tiled floor flattens, the screens on the ticket machines go blank,
    and the burning wreck in the clearing is left as geometry alone.}
    \label{fig:whitemodel}
\end{figure}

\paragraph{White model.} \engine can render a sequence a second time as a \emph{white model}, with
geometry and camera motion preserved and surface appearance replaced by a neutral material and its
shading. This is the coarse input a coarse-to-real generator is trained against, where the structure
is given and the appearance is what the model has to supply. The whitening happens in the renderer
rather than through edits to individual scene components. The engine's default-material show flag
covers
buildings, terrain and its procedural grass, foliage and instanced meshes, HLOD proxies, spline
geometry such as rails, and buildings streamed in at runtime. Replacing materials component by
component in an earlier implementation reached mesh components but missed landscape grass, level
instances and HLOD buildings. The white-model render uses the same level sequence and camera
configuration as its textured counterpart (\Cref{fig:whitemodel}). Because the show flag replaces
surface albedo but not illumination, the result is desaturated after rendering to obtain a neutral
grey image while retaining shading. The sequence manifest states whether a white-model render is
included.

\subsection{Characters}

\begin{figure}[H]
    \centering
    \begin{subfigure}{0.495\textwidth}
      \includegraphics[width=\linewidth]{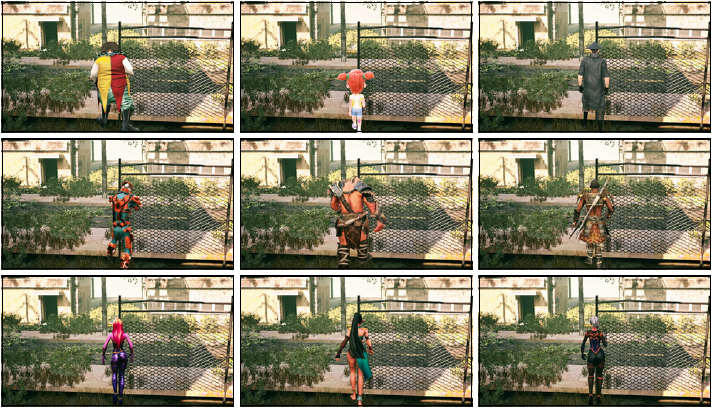}
      \caption{Humanoids}
    \end{subfigure}\hfill
    \begin{subfigure}{0.495\textwidth}
      \includegraphics[width=\linewidth]{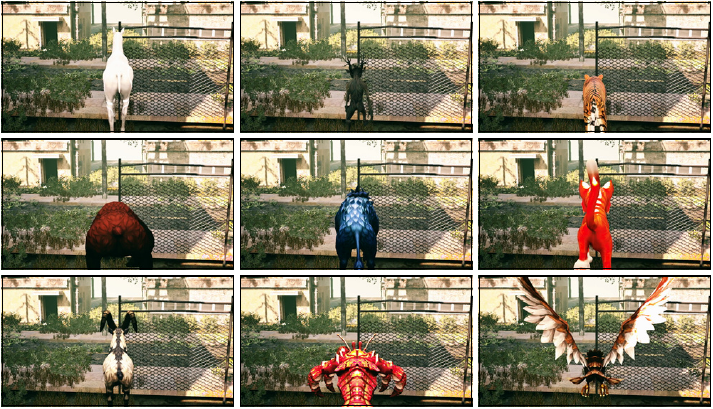}
      \caption{Animals and creatures}
    \end{subfigure}
    \caption{\textbf{Character library.} Asset-review sheets --- the avatars shown in one shared scene at a
    fixed pose --- rather than frames from the released dataset.
    Third-person sequences are rendered with humanoids, quadrupeds and birds, which differ in
    body plan, gait and how they turn, so subject articulation is a diversity axis in its own
    right.}
    \label{fig:charlib}
\end{figure}

The \engine asset library contains 70+ animated characters, including humanoids
alongside animals and creatures (\Cref{fig:charlib}). Body plans differ: bipeds pivot in place, quadrupeds and birds turn along arcs, and each is
authored at its native scale and ground contact. For a dataset aimed at 4D reconstruction of
deforming subjects, this variety in articulation and gait is as important as scene variety.

\subsection{Scalability and statistics}
\label{sec:stats}

\begin{table}[htbp]
\centering
\tablestyle{5pt}{1.2}
\caption{\textbf{\engine capacity and the current \data release.} The upper block describes the
engine and asset library, not the coverage of every released subset. In the lower block, scenes are
counted per viewpoint and therefore do not add; 32 is the number of distinct environments.}
\label{tab:stats}
\begin{tabular}{lrrrrr}
\toprule
\multicolumn{6}{l}{\textbf{Engine/library capacity}} \\
\midrule
\multicolumn{2}{l}{Environments}    & \multicolumn{4}{l}{30+ artist-built UE scenes, interior to city scale} \\
\multicolumn{2}{l}{Lighting states} & \multicolumn{4}{l}{day, dawn, evening, sunset, night} \\
\multicolumn{2}{l}{Weather states}  & \multicolumn{4}{l}{clear, overcast, fog, snow} \\
\multicolumn{2}{l}{Materials}       & \multicolumn{4}{l}{textured render or untextured white model} \\
\multicolumn{2}{l}{Characters}      & \multicolumn{4}{l}{70+ animated humanoids, animals and creatures} \\
\multicolumn{2}{l}{Routes}          & \multicolumn{4}{l}{navmesh planner, reactive explorer, manual recording} \\
\multicolumn{2}{l}{Scales by}       & \multicolumn{4}{l}{scene $\times$ state $\times$ character $\times$ route $\times$ viewpoint} \\
\midrule
\multicolumn{6}{l}{\textbf{Current release}} \\
\midrule
& \textbf{Scenes} & \textbf{Sequences} & \textbf{Frames} & \textbf{Duration} & \textbf{Volume} \\
\midrule
First-person   & 23 & 2,910 & 10.8\,M & 100.5\,h & 5.9\,TB \\
Third-person   &  7 & 1,885 &  8.3\,M &  76.5\,h & 4.4\,TB \\
$360^\circ$ panoramic & 15 & 1,208 & 2.8\,M & 25.8\,h & 8.4\,TB \\
\cmidrule(lr){1-6}
Total          & 32 & 6,003 & 21.9\,M & 202.7\,h & 18.7\,TB \\
\bottomrule
\end{tabular}
\end{table}

The current release contains 6,003 sequences from 32 environments, totalling 21.9M rendered frames,
202.7 hours of video and 18.65\,TB of data. \Cref{tab:stats} reports the breakdown by viewpoint.
The name \data refers to the 10.8M first-person frames; third-person and panoramic sequences add
further observations of routes defined in the first-person trajectory set.

\begin{figure}[htbp]
    \centering
    \includegraphics[width=\linewidth]{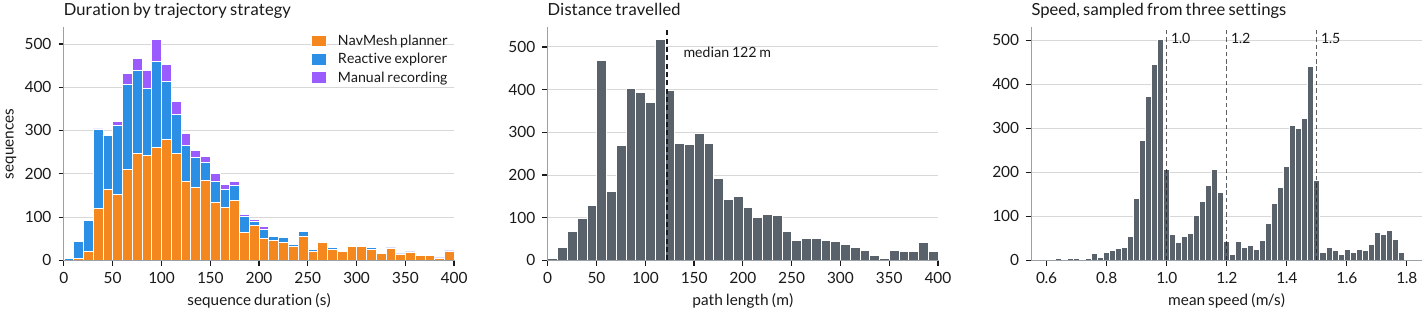}
    \caption{\textbf{Distributions over the released sequences} ($n=6003$, from per-sequence
    metadata). Left: sequence duration, stacked by trajectory strategy in the colours of
    \Cref{fig:traj} --- reactive routes end sooner than planned ones, which run on to the long tail.
    Centre: distance travelled, a median of $122$\,m. Right: mean speed per sequence, which falls
    into three clusters because the motion model samples one of three walking speeds; each cluster
    sits a little below its setting, the cost of turning and rate limiting.}
    \label{fig:stats}
\end{figure}

All three trajectory strategies are represented: 3,662 sequences from the navmesh planner, 1,996
from the reactive explorer and 345 from waypoint trajectories. Median durations are 115\,s for
planner routes, 105\,s for waypoint trajectories and 81\,s for
reactive ones --- while their speed is drawn from three settings, $1.0$, $1.2$ and
$1.5$\,m/s, so its distribution is three clusters and not one (\Cref{fig:stats}).
By scene category the largest share is urban, at 2,128 sequences, followed by town and period
settings at 1,668, interiors at 639, historical and fantasy at 633, and cyberpunk at 603.

\section{Related Work}
\label{sec:related}

\subsection{Simulators and engine interfaces}

A simulator advances an agent through a world and returns observations; an engine interface exposes
the controls and state needed to do so. Photorealistic
simulators are built on Unreal, on Unity~\citep{kolve2017ai2thor} and on platforms such as Isaac
Sim~\citep{nvidia_isaacsim}; a parallel line is organised around physical interaction, hosting
articulated objects for manipulation~\citep{xiang2020sapien}. On UE, UnrealCV~\citep{qiu2016unrealcv}
opened the engine to computer vision and UnrealZoo~\citep{zhong2025unrealzoo} extends it with over a
hundred worlds; AirSim~\citep{shah2018airsim} and CARLA~\citep{dosovitskiy2017carla} built
domain-specific simulators for aerial and driving research; the Habitat
line~\citep{puig2023habitat3} did the same for indoor embodied
agents; and AirSim360~\citep{ge2025airsim360} extends the idea to omnidirectional drone view, with
equirectangular RGB, slant-range depth and automated trajectories. Many of these systems expose a
curated Python API and are distributed with a particular simulator project, which can limit access
to engine functionality or complicate the use of third-party UE content. SPEAR~\citep{spear2026}
instead uses UE runtime reflection and attaches to an existing project as plugins. \engine adopts
that interface and extends it into an offline production pipeline with asset preparation, trajectory
construction, Movie Render Queue rendering, streaming post-processing, parallel scheduling and
recovery.

\subsection{Synthetic video datasets}

\paragraph{Rendered from existing 3D assets.} Several synthetic video datasets are rendered from
existing 3D assets. The BEDLAM
line~\citep{black2023bedlam,tesch2025bedlam2} renders clothed human motion at scale, and is an asset
source for later work. Dynamic Replica~\citep{karaev2023dynamicstereo} renders thirteen animated
character models for long-term correspondence.
PointOdyssey~\citep{zheng2023pointodyssey} renders animated assets outside a game engine,
specifically for long-range tracking: $104$ videos averaging $2{,}000$ frames, with deformable
characters driven by real motion capture, and with characters, materials, lighting and assets
randomised across sequences. Syn4D~\citep{syn4d} composes store-sourced UE environments with
animated objects and BEDLAM humans, and renders them into multi-view video with depth, dense tracks
and parametric human pose; its stated aim is that any pixel can be unprojected to 3D at any time and
from any camera. Its supervision is closely related to ours. Syn4D uses general multiview camera
patterns, whereas \method defines first-person, third-person and panoramic
viewpoints with distinct traversal semantics.
OmniX~\citep{omnix2026} renders $80$K scenes into $1.28$M multi-view videos from a UE5 data engine,
with dynamic objects composited into static environments; it
releases no action stream.

\paragraph{Recorded from video games.} A second group records a finished game instead of
rendering one --- a technique going back to intercepting a game's traffic to its graphics
hardware~\citep{richter2016playing}, extended in turn to depth, cameras and instance
segmentation~\citep{richter2017playing} and to body joints~\citep{fabbri2018learning}. OmniWorld~\citep{zhou2025omniworld} takes one of its four domains this way, lifting synchronised RGB
and depth off the screen with a shader injector and a recorder; most of its $300$M+ frames come
from curated public datasets. Sekai's game
subset~\citep{sekai2025} records a UE5 landscape-photography title with a screen recorder, logging
location, weather and camera trajectory as it plays, alongside a far larger subset of real video.
The Generative World Renderer~\citep{genworldrenderer} captures $4$M
frames of RGB synchronised with five G-buffer channels, for inverse rendering and
G-buffer-conditioned synthesis. WildWorld~\citep{wildworld} records an action role-playing game as
$108$M frames with skeletons, world state, camera pose and depth, under a vocabulary of over $450$
actions, compared with the four quantized axes in the \data trajectory-derived action stream.
Gameplay recorded together with
the control input that drove it is by now an established source of action-conditioned
video~\citep{che2025gamegen,guo2025mineworld,yu2025gamefactory,kanervisto2025world}. What can be
annotated this way is decided by the capture hook, which may intercept graphics calls, record the
screen, combine synchronized displays, inject a shader or read game memory.
The available labels therefore differ by title and by the quantities exposed through each capture
method.

\section{Applications}
\label{sec:applications}

\paragraph{Interactive world models.} RGB, rendered camera trajectories and trajectory-derived
actions provide conditioning signals without a separate pose-estimation step
(\Cref{sec:trajfiles}). Third-person sequences distinguish camera motion from character motion,
while matched perspective and panoramic observations provide reference content outside the
perspective field of view.

\paragraph{Dynamic 4D reconstruction.} Dense optical flow and long-range point tracks provide
short- and long-range correspondence, including 3D position, 2D projection and visibility through
occlusion (\Cref{sec:tracks}). Separate camera and character trajectories help distinguish camera
motion from the motion of an articulated subject.

\paragraph{Generative rendering.} White-model and textured pairs provide frame-aligned structural
control for coarse-to-real generation~\citep{coarse2real}. Environmental variants provide further
pairs in which geometry and motion are
fixed while appearance changes (\Cref{sec:lightingstates}).

\section{Conclusion and Future Work}
\label{sec:conclusion}

We presented \method, comprising \engine, a pipeline for rendering trajectories through
artist-built UE environments, and \data, the resulting release. The dataset name refers to 10.8M
first-person frames; including third-person and panoramic observations, the current release contains
21.9M frames in 6,003 sequences. Its task-oriented subsets provide
combinations of metric depth, optical flow, long-range 2D/3D point tracks, calibrated cameras and
trajectory-derived actions, together with first-person, third-person and panoramic RGB observations.

Each released annotation has a specified relationship to the pixels it accompanies. Raster
annotations share a rendered instant;
camera calibration describes the camera that produced the frame; a long-range track is projected
through that camera and its visibility decided against the rendered depth. Each sequence declares
which outputs it carries. \engine composes environments, routes, characters, viewpoints and
environmental states, while \data denotes the sequences released from those configurations.

\label{sec:limitations}\label{sec:roadmap}

The current system has three main limitations. \textbf{(1)} Each third-person sequence contains one
independently moving, trajectory-annotated character; other dynamic scene elements are not separately
controlled or annotated. The follow camera also rides several metres behind that character, where the
route provides no traversability guarantee. In a cluttered scene it can enter geometry,
and the sequence is withheld. Populating a sequence with several tracked characters, and making the
follow camera occlusion-aware, would address both. \textbf{(2)} Dynamic global illumination and
artist-built assets do not model sensor effects; noise, rolling shutter and lens distortion are
absent from the rendered frames.
\textbf{(3)} This report describes the pipeline and the data it produces, but reports no downstream
results. The next step is to train on the released data and measure the result on standard
benchmarks.


\bibliographystyle{abbrv}
\bibliography{references_back}

\end{document}